\documentclass{article} 
\usepackage{collas2022_conference}
\usepackage{times}
\usepackage{epsfig}
\usepackage{amsmath}
\usepackage{amssymb}
\usepackage{dirtytalk}
\usepackage{subcaption}
\usepackage{graphicx}
\usepackage{float}
\usepackage{textcomp}
\usepackage{todonotes}
\usepackage{multicol}

\usepackage[pagebackref=true,breaklinks=true,colorlinks,bookmarks=false]{hyperref}

\usepackage{amsmath,amsfonts,bm}

\def\eqref#1{equation~\ref{#1}}

\def\1{\bm{1}}

\DeclareMathAlphabet{\mathsfit}{\encodingdefault}{\sfdefault}{m}{sl}
\SetMathAlphabet{\mathsfit}{bold}{\encodingdefault}{\sfdefault}{bx}{n}

\usepackage{hyperref}
\hypersetup{
    colorlinks=true,
    linkcolor=red,
    filecolor=magenta,      
    urlcolor=blue,
    citecolor=purple,
    pdftitle={Overleaf Example},
    pdfpagemode=FullScreen,
    }

\newcommand{\ClassesDrift}{Virtual concept drift }
\newcommand{\classesDrifts}{ virtual concept drifts }
\newcommand{\ClassesDrifts}{Virtual concept drifts }

\newcommand{\InstanceDrifts}{Domain drifts }
\newcommand{\instanceDrift}{ domain drift }
\newcommand{\InstanceDrift}{Domain drift }
\usepackage{authblk}

\newcommand{\modif}[1]{\textcolor{black}{#1}}

\title{Understanding Continual Learning Settings with Data Distribution Drift Analysis}

\author{Timoth\'ee Lesort}
\author{Massimo Caccia}
\author{Irina Rish}
\affil{Université de Montréal, MILA - Quebec AI Institute}
\date{}

\collasfinalcopy

\begin{document}

\maketitle

\begin{abstract}
Classical machine learning algorithms often assume that the data are drawn i.i.d.  from a  stationary probability distribution. Recently,  {\em continual learning} emerged as a rapidly growing area of machine learning where this assumption is relaxed, i.e. where the data distribution is non-stationary and changes over time.
\modif{This paper represents the state of data distribution by a context variable $c$. A drift in $c$ leads to a data distribution drift.
 A context drift may change the target distribution, the input distribution, or both. Moreover, distribution drifts might be abrupt or gradual. In continual learning, context drifts may interfere with the learning process and erase previously learned knowledge; thus, continual learning algorithms must include specialized mechanisms to deal with such drifts. }
In this paper, we aim to identify and categorize different types of \modif{context drifts} and potential assumptions about them, to better characterize various continual-learning scenarios. 
Moreover, we propose to use the distribution drift framework to provide more precise definitions of several terms commonly used in the continual learning field.
\end{abstract}

\section{Introduction}
 Continual learning  (CL) involves the ability to learn from a  non-stationary data stream and accumulate knowledge over time, while successfully dealing with the {\em catastrophic forgetting} problem \citep{French99}, i.e. the 
 interference between the new knowledge and the previously learned one.

CL algorithms are usually classified based on the type of \say{memory} mechanisms they use, including   replay   \citep{lesort2020continual,Chaudhry19, Aljundi2019Online,Belouadah2018DeeSIL,lesort2018generative,wu2019large,Hou_2019_CVPR,caccia2019online}, dynamic architecture \citep{fernando2017pathnet,Rusu16progressive,Li17learning} and regularization   \citep{kirkpatrick2017overcoming,Zenke17,Ritter18Online, schwarz2018progress,lesort2019regularization}. However, it is not always clear how each specific approach can benefit a particular continual learning scenario.
In this paper, we start by assuming that catastrophic forgetting (or learning interference) is caused by changes in the data distribution or changes in the learning criterion. We will refer to these changes as drifts.
There are multiple types of drifts, and it is unlikely that any specific approach can work well for all of them. 
This hypothesis looks reasonable since the CL problem is NP-Hard  \citep{knoblauch2020optimal}. 
We could therefore characterize CL algorithms by their ability to learn under specific types of distribution drifts, associated with different CL settings.

\begin{figure}[h]
    \centering
    \includegraphics[width=0.75\linewidth]{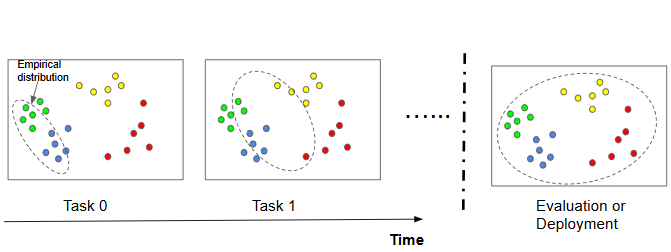}
    \caption{Example of a continual learning scenario. The learning algorithm can only access the empirical data distribution determined by a set of training samples. The empirical distribution  drifts continually, allowing to explore the full underlying data distribution. The goal of continual learning is to produce a model that generalizes well  to  this full data distribution.}
    \label{fig:virtual_drifts}
\end{figure}

The paper is organized as follows.    In Sec.~\ref{sec:related}, we give a brief overview of other fields dealing with context drifts.
In Sec.~\ref{sec:drifts}, we discuss how to use the various types of context drifts to describe non-iid characteristics of continual learning problems.
Furthermore,  in Sec.~\ref{sec:perf_eval} we discuss evaluation criteria for  continual learning algorithms.
Next, in Sect.~\ref{sec:assumptions}, we present different types of assumptions about the distribution drifts that can be used to improve CL algorithms.  
In Sec. \ref{sec:terminology}, we benefit from the drifts types to propose a detailed description for incremental learning and lifelong learning.
Moreover,  in Sec. \ref{sec:discuss} we discuss the benefits and limitations of meta-learning approaches for continual learning and implicit supervision that can be provided for learning algorithms.
Finally, we conclude by summarizing the benefits of the proposed description of continual-learning scenarios.


\section{Related work}
\label{sec:related}

The topic of distribution drifts (or shifts)  is well-studied in machine learning and statistics. 
For example,  the field of {\em out-of-distribution (OOD) generalization} aims at developing learning algorithms able to perform well on test distributions that are different from training distribution \citep{arjovsky2019invariant, ahuja2020invariant, krueger2021outofdistribution}. A typical assumption is the existence of some  invariant (causal) mechanism underlying the data distribution, besides variations  due to other factors; a common approach is to train a model on multiple datasets, or  \say{environments}, in order to  learn a robust predictor based on invariant rather than spurious features.

Other closely related fields include transfer learning \citep{weiss2016survey} and domain adaptation \citep{NIPS2006_b1b0432c, wang2018deep}, which aim at adapting a  model trained on one distribution, or domain, to a new data distribution, but without necessarily remembering the past distribution. For example, in robotics, many approaches are focused on  transferring a model trained in a simulated environment to real environments \citep{peng2018sim, Traore19DisCoRL, james2019sim}.

Continual learning is a more challenging problem since the goal here is not only to adapt to new data distributions but also to avoid interference with already acquired knowledge from past data distributions \citep{Lesort2019Continual}. To tackle this challenge across a wide range of problem settings  in a  robust and scalable way, 
continual learning approaches can borrow ideas and techniques from other research domains such as few-shot learning \citep{Lake11, Fei-Fei06}, transfer learning \citep{Pratt93,Zhao17}, knowledge distillation \citep{hinton2015distilling}, meta-learning \citep{Brazdi2008Metalearning,hospedales2020meta}, learning from sparse supervision and anomaly/concept-drift/out-of-distribution detection \citep{gama2014survey}. Initially, the goal of continual learning is to learn from a sequence of varying datasets without forgetting the previously learned knowledge and continuously improving.
In a broader perspective, continual learning algorithms could ultimately build versatile, "broad" AI agents (as opposed to the current "narrow" AI), through a potentially infinite sequence of tasks. Then, contributing to the development of general artificially intelligent agents could be also an end goal of continual learning.  


\modif{Other works on continual learning proposed to frame the various scenarios of continual learning. \cite{van2019three} proposed task-incremental, class-incremental, and domain-incremental scenarios depending on the evolution of data distribution and the availability of the task label at test time. \cite{de2021continual} proposed the data-incremental scenario with no notion of tasks, i.e., no latent variable that could explain the evolution of the data distribution. On the other hand, \cite{caccia2019online} proposed a comprehensive framework to characterize the various learning paradigm such as continual learning, meta-learning, continual meta-learning, and meta-continual learning. Our paper focuses on the evolution of data distribution. We also build upon the distribution shift literature, notably \cite{kelly1999impact,vzliobaite2010learning, gama2014survey}.   Moreover, we detail all the possible context drifts in existing scenarios and how they impact differently continual learning algorithms.}

\section{Context drifts in continual learning}
\label{sec:drifts}
A clear characterization of the data stream evolution  is necessary to better understand the capabilities and limitations of continual learning approaches. We propose an attempt at such characterization.

\subsection{Continual learning and context drift}

We introduce here a hidden context variable $C$ that  determines the current state of the  data distribution, similarly to \citep{caccia2020online}.  Let $\mathbb{C}$ denote the set of all possible contexts.
We use the change, or drift, in the context variable to characterize the evolution of the data distribution. Hence, a context drift is equivalent to a data distribution drift. 
In this section, we describe different types of context drift in continual learning scenarios.

Given the context $c$, data can be sampled i.i.d.\ from $P(X=x,Y=y|C=c)$.
In a supervised setting, $x$ is the input sample and $y$ is the class label. In reinforcement learning (RL), $y$ can be viewed as the action taken by an agent. 
Changes in the data distribution are assumed to be caused by changes in the context variable, i.e. by the \textit{context drift} (not to be confused
with the \textit{concept drift}, a specific type of context drift   discussed in section \ref{sub:drifts}). In other words, the data non-stationarity is caused by a hidden stochastic process $\{C_t\}_{t=1}^T$, where $C_t$ is the context variable at time $t$.

In many continual settings, a period within which a context remains fixed  (corresponding to a stationary data distribution) is called a  {\em task},  and usually corresponds to a  learning experience determined by some specific learning purpose \citep{Lesort2019Continual}. Switching from one task to another 
is usually associated with a change in the learning process, involving a new context or a new learning objective.

In the next section, we present a characterization of context drifts.

\subsection{Types of drifts}
\label{sub:drifts}
As it was mentioned above, a \textit{context drift} is a change in the context variable, associated with a change in the data distribution; it is similar to \say{population drifts} \citep{kelly1999impact}. We distinguish between the  two kinds of context drifts that we refer to as real concept drifts and virtual  drifts \citep{gama2014survey, gepperth2016incremental}.

 As said in \cite{gama2014survey} the \emph{concept} is the relation between input data and target data.
 The concept is the sense that $y$ gives to $x$. In this paper, the concept of a label $y$ is represented by the data $x$ associated with this label. 

\begin{figure}[H]
 
    \centering
    \begin{subfigure}[H]{0.4\linewidth}
        \includegraphics[width=0.9\linewidth]{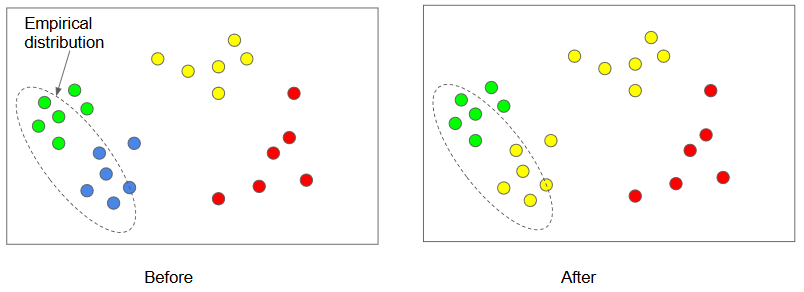}
        \caption{\textbf{Real Concept Drift}}
    \end{subfigure}
    \centering
    \begin{subfigure}[H]{0.4\linewidth}
        \includegraphics[width=0.9\linewidth]{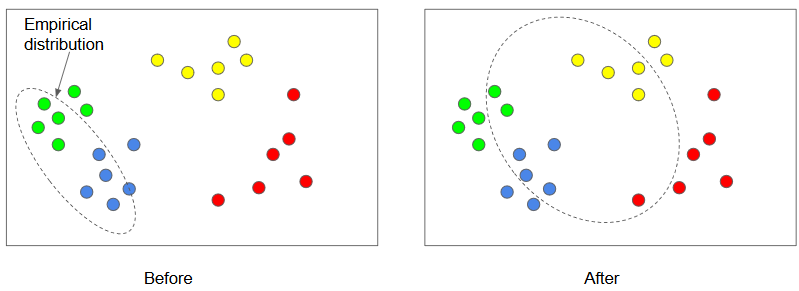}
        \caption{\textbf{Virtual Drift}}
    \end{subfigure}
    \caption{Real concept drift vs Virtual drift. Dots are data point $x$ while the colors correspond to a associate $y$. Real Concept Drift: the label distribution $P(y|x)$ changes, while the empirical distribution $P(x)$ do not change. Virtual Drift: the empirical distribution $P(x)$ changes but not the label distribution $P(y|x)$. Same-color points belong to the same class. }
        \label{fig:drifts}
\end{figure}

\medskip

\textbf{Real concept drifts:} \textbf{drift in $P(y|x)$ with $P(x)$ fixed.} \quad Real concept drifts \citep{gama2014survey, gepperth2016incremental} happen when the relation between the input and the target changes (Fig \ref{fig:drifts}.a). Specifically, $P_t(y|x) \neq P_{t+1}(y|x)$ whilst $P_t(x) = P_{t+1}(x)$, with $P_t(y|x)$ being the probability $P(Y=y|X=x)$ at time $t$. 
For example in recommendation systems, a user's mood (the mood is here the context variable $t$) could determine its preference  (the labels $y$). Then a change in $t$ might lead to a real concept drift in $y$. 

 \textbf{Virtual drift \citep{gama2014survey}: Drift in $P(x)$ without affecting $P(y|x)$}. In this case, the variation in the data distribution does not affect the relation between input $x$ and labels $y$ (Fig \ref{fig:drifts}.b). This drift is also known as covariate shift \citep{Suhiyama2012Machine} or dataset shift \citep{Quionero2009Dataset,MORENOTORRES2012521}. 
We distinguish two specific kinds of virtual drifts:
\begin{itemize}
    \item \textbf{\ClassesDrift: Drift in $P(y)$ without affecting $P(y|x)$}: Data points sampled after the drift have  new labels. In supervised learning, it involves observing new data from new classes $y$. 
 Specifically, the joint distribution  $P_t(y,x)$ shifts through $P_t(y) \neq P_{t+1}(y)$ while $P_t(x|y) = P_{t+1}(x|y)$. This type of drift is also known as \say{label shift} \citep{zhang2020dive}.
  \item \textbf{\InstanceDrift: Drift in $P(x)$ without affecting $P(y|x)$ nor $P(y)$}: Data points sampled after the drift are new, i.e. new data of the same label space. 
\end{itemize}

 \medskip
 
 \begin{figure}[H]
 
    \centering
    \begin{subfigure}{0.4\linewidth}
        \includegraphics[width=0.9\linewidth]{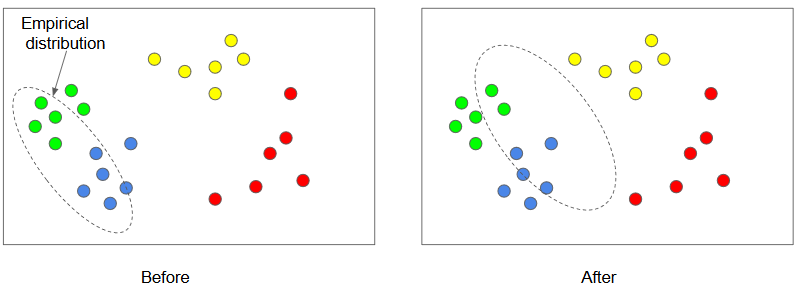}
        \caption{\textbf{\InstanceDrift}}
    \end{subfigure}
    \begin{subfigure}{0.4\linewidth}
        \includegraphics[width=0.9\linewidth]{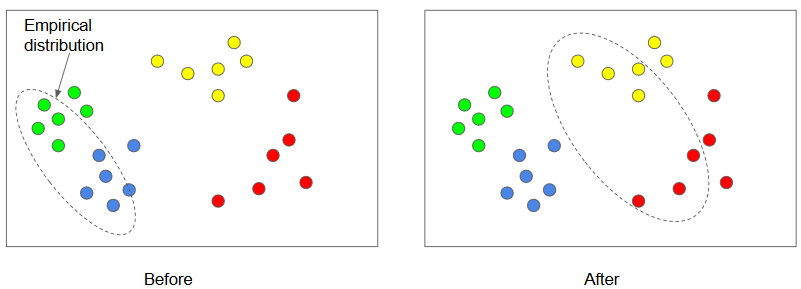}
        \caption{\textbf{\ClassesDrifts}}
    \end{subfigure}
        \caption{Virtual drifts special cases. \InstanceDrifts: new samples are from known class(es). \ClassesDrifts: new samples are from unknown class(es). In the figures, points with same colors are point within a same class.}
    \label{fig:virtual_drifts}
\end{figure}
 
We can think of a new type of drifts in the learning process which is the \textit{criterion drift}: The data does not change but the learning criterion does: e.g. the loss. 
For example, in reinforcement learning, algorithms have to maximize their accumulated rewards. For this, a classic strategy is an exploration-exploitation strategy. We can assume that the loss for exploitation and the loss for exploration are two different losses. Therefore, while the algorithm learns how to explore, it should not forget what has been learned while it was exploiting some reward sources. Changes in the loss function (or in the weights of loss function) might lead to catastrophic forgetting and should therefore be addressed in an adapted way.
However, \textit{criterion drift} and \textit{real concept drift} affect the learning process in the same way, i.e. the expected behaviour changes. Hence, algorithms might deal with both in the same way. The difference is that criterion drifts are caused by the algorithm's choices and not by changes in the data distribution (as real concept drift). Learning algorithms should then be always aware of criterion drifts.

\medskip

\modif{Any drift in $p(x,y)$ might also be described as a composition of the drifts described in this section}. Nevertheless, it is fundamental to distinguish the different types of drifts since they will not affect the learning process in the same way. We detailed the differences it involves in section \ref{sec:terminology}. \modif{The idea would be able to deal with elementary data drifts independently to be able to target any composition of drifts}.
Besides the drifts type, the annotation or the automatic detection of drift is also crucial for CL algorithms to enable a memorization process at the right time.

\subsection{Drift Annotations}
\label{sub:drift_annotation}

In classical machine learning, learning paradigms are dependent on the type of supervision available, mainly, supervised learning relies on label supervision, reinforcement learning on reward supervision. 

In continual learning, algorithms do not \textit{only} rely on those types of supervision but also on drifts categorization \citep{Lesort2019Continual}. 
Indeed, context drift might be annotated by \say{drift labels} to help algorithms to prevent interferences. 

In continual learning-benchmarks, the drift labels often give the task boundaries or task index. 
It is referred to as \say{task labels} 
. In the continual-learning literature, we find several specific scenarios based on the availability of task labels \citep{Lesort2019Continual}.
Hence, the tasks index might be available (1) for training, (2) for training and testing or  (3) not provided at all. The cases (1) and (2) have been associated to, respectively, \say{task-incremental learning} and \say{domain-incremental learning} (or \say{class-incremental learning}) in \citep{van2019three}. The last one has also been associated to \say{task-agnostic learning}. 

\modif{It is important to note the task label 
is different from the context variable $c$.  $c$ is a theoretical latent variable that represents the data distribution state. Hence, a task label is only a piece of information about $c$ or about $c$'s evolution.}
With tasks labels can come also supplementary information such as the type of drift, the number of new labels, or other a priori knowledge or meta-knowledge \citep{Douillard2020Insights} about the new task.
On the other hand, the learning scenario does not always have clear task boundaries, so a more general hypothesis is to suppose that the drift labels might signal variations in the data distribution and not necessarily a clear task transitions. 


In this paper, we wish to define types of scenarios based on the context drifts and not on the availability of task labels. The task labels availability only changes how difficult the task is. \modif{Similar to classification, where the more data labels are available, the easiest the learning task is.}

\section{Evaluation Protocols in Continual Learning scenarios}
\label{sec:perf_eval}

The characterization of data distributions in continual learning is a clear starting point to describe learning scenario characteristics. However, the evaluation process is also important to describe what are the expectations for a given approach.
In this section, we describe in a general way continual learning scenarios and associated evaluation protocols.

\subsection{Evaluation protocols: Cumulative and Current Performance}

The evaluation protocol defines what is expected from a continual-learning algorithm: its goal. Algorithms learn to optimize their evaluation protocols. 
We again assume a stream of contexts (or tasks) $\{C_t\}_{t=1}^T$ which the learner is exposed to.

 In continual learning, we distinguish two types of evaluation: maximizing the performance of the model's final state or maximizing the performance while learning. The first is the most used performance evaluation in continual learning often referred to as \say{final performance} \citep{Chaudhry19, kirkpatrick2017overcoming,shin2017continual,lesort2018marginal}, however in a more general scope, it evaluates the \textbf{current performance} of an algorithm on a test task at a particular point in time.
\begin{equation}
        \sum_{\modif{t=1}}^T \mathbb{E}_{(x,y) \sim P(x,y|c_t)}[\ell(f_{\theta_T}(y|x))],
\end{equation}
where $\ell(.)$ is the criterion and $f_{\theta_T}(.)$ is the function learned at the end of the sequence, i.e. in the final state $t=T$.

 The second measures the algorithms \textbf{cumulative performance} throughout its \emph{life}, it is referred to as \textit{Online Cumulative Performance} \citep{caccia2020online}:
 \begin{equation}
    Cumul_{\{C_t\}_{t=1}^{T}} = \sum_{\modif{t=1}}^T \sum_{(x,y) \sim P(x,y|c_t)} \ell(f_{\theta_{\modif{t-1}}}(y|x)).
\end{equation}
It evaluates the performance of \say{learning while doing} where the goal is the results achieved while learning, i.e. while $\theta$ changes and not with a fixed $\theta$ as for current/final performance evaluation. 

The expected future cumulative performance might be maximized in a meta-learning approach. The meta-algorithm goal is to maximize cumulative performance for the next task sequence, i.e. maximize $\mathbb{E}[Cumul_{\{C_t\}_{t=T}^{T+T'}}]$, with $T'$ the length of the future sequence.

\medskip

\modif{\textit{\textbf{Current performance} measures the performance of the current model via a test evaluation, while \textbf{cumulative performance} measures the performance realized by the model while learning on training data.}}

\medskip

Those two evaluation protocols are used for different kinds of applications. 
The cumulative performance is mainly suited for algorithms that continue to learn while deployment, e.g. reinforcement learning agents that continuously improve at their tasks. 
The current performance evaluates the quality of a given model with fixed weights. It is more suited for use cases where models are deployed \emph{frozen} or when models expect external validation before deployment. 
\modif{In a reinforcement learning setting, cumulative performance measures the amount of reward accumulated during training, while the current performance measures the accumulated reward in a test task after training.}
To integrate the evolution of current performance at different steps of the training process, we can average several \say{current performance} from different points of time as in \citep{rebuffi2017icarl}. 

\medskip

The evaluation protocol chosen for an algorithm defines its objective, either by evaluating the performance while learning or the final performance.
In the next section, we detailed the specific types of drifts that can happen in training data distribution. 
We will also describe the assumptions that could help algorithms to deal with those drifts.

\section{Drift Assumptions}
\label{sec:assumptions}

To ease the learning process, reduce the solution space, or reduce the need for supervision, algorithms may rely on assumptions about data. 
In continual learning, we may assume the characteristics of context drift. For example, we can assume that certain types of drifts described in Sec \ref{sec:drifts} do not happen locally or permanently, or we can assume some underlying patterns in drifts or drifts characteristics.

\subsection{Stationarities}

Stationarity aspects of the learning process, for example in the context variable. Hence, we may assume local stationarities or general stationarities. In this section, we describe common assumptions that might hold in a continual-learning environment.

\medskip

\textbf{Concept stationarity} \quad The number of labels (e.g. classes, environment...) is fixed by the designer at the beginning and it does not change through time. This is akin to domain-incremental learning \citep{van2019three}.
\underline{Example:} An urban camera that detects bicycles and continually improves. There is no other labels than bicycle vs no-bicycle in new data.

\textbf{Target/Reward stationarity} \quad  In this case, we consider that the label associated with a data point $x$ is fixed, i.e. $P(x|y)$ (or $P(y|x)$) is static.
\underline{Example:} In many classification scenarios, the data annotation never changes.

\textbf{Finite World} \quad The data's possible instances are fixed and finite. Therefore at a certain point, all instances possible have been seen.
\underline{Example:} Board games.


\textbf{Criterion stationarity} \quad  The objective function is constant. 
\underline{Example:} A camera that detects bikes has a fixed objective even if the input distribution changes through time.



\textbf{Non-backtracking chain} \quad  The non-backtracking chain assumption means that we assume that everything is learned just once and will never append again in the training process. This assumption is applicable in most continual learning benchmarks, e.g., in disjoint scenarios (class-incremental). The non-backtracking chain helps to evaluate which task has been forgotten or not. However, this assumption is rarely true in real scenarios, therefore it should not be exploited by approaches in continual learning benchmarks. An example of continual learning setting where the non-backtracking chain does not apply is OSAKA \citep{caccia2020online}.\\ 

\medskip

The stationarities are the simplest (and the strongest) assumptions, in some cases, we can also make assumptions on drift patterns.

\subsection{Drift Patterns and Assumptions}

Drift in the data distribution can be subject to patterns, for example, variation in traffic prediction might depend on the hour and the date. Those patterns can be incorporated into continual-learning approaches to improves results. The simplest pattern is to assume no drifts, e.g. no real concept drift, either locally or permanently.

\quad We call a data stream $\alpha$-locally-invariant when $P(C_t \!=c | C_{t-1}\!=c)=\alpha$ \citep{caccia2020online}. Namely, the data distribution is locally stationary, i.e. stationary within a local time window. 
Those local stationarities are assumed in most continual learning benchmarks, they help to train models but also to evaluate them by keeping drifts sparse.

\modif{The context set can also be assumed finite, i.e., the set $\mathbb{C}$ containing all contexts is finite. This assumption is denoted \say{closed world assumption} \citep{mundt2020wholistic}.}
 At some point in time, all the contexts are explored and all drifts will lead to context already explored.
For example, a robot that explores a building. At some point in time, the robot will have explored all the rooms and will never find new rooms or places in the building.
This assumption is useful when a model is deployed static, we assume that contexts have all been explored sufficiently while training the algorithms can make good predictions on future observations. By sufficiently, we mean that if a new context happens at test time the model can generalize to it without retraining.
The opposite of a closed world is an open-ended world \citep{Doncieux2018Open} where a \textit{never-ending drift} might happen with infinite unexplored contexts. 

A last type of pattern is the recurrent concept drift \citep{gama2014survey}. In this case, the drifts obey some kinds of cyclic patterns. Continual learning approaches may try to take advantage of them to predict future contexts and improve algorithms' performance as in the traffic example of the introduction.

\subsection{Drift Intensity}

Drifts might happen with various intensities, suddenly or smoothly.
The intensity of drift is then characterized by the variation in the context variable in a short period. 
\modif{Drifts might be automatically detected, e.g., when drifts are abrupt, which lowers the need for drift annotation. Still, the annotation can also be needed to disentangle noisy variations from a significant drift.
For example, the drift smoothness might be an essential parameter to consider to detect drifts successfully.} 
Hence, the assumptions on the intensity of drifts might help to disentangle context drifts from spurious variations in data.
Moreover, the memory mechanisms might be adapted to the abruptness of the drifts to take advantage of specific settings.

\medskip

To summarize this section, in order to handle efficiently the different types of scenarios, algorithms may formulate assumptions on their drifts.
The assumptions might be seen as a weakness of approaches since it makes them dependant on conditions. However, assumptions help to reduce the search space and ease the learning process. They may help to reduce the need for supervision and make unsolvable problems solvable. 
Nevertheless, it is important to describe clearly assumptions needed to train an algorithm to understand the approach and potential transfer to other scenarios.

\section{Continual Learning Terminology and Real-Life Scenarios}
\label{sec:terminology}

\modif{The continual learning research field aims at creating learning tools for real-life applications.
As there is a high variability of potential applications, we can not create benchmarks for each of them. It is then necessary to define and disentangle the qualities required to overcome a given application and be able to assess them. The goal would be to have a set of benchmarks that assess the algorithm's performances on a set of criteria to transfer those algorithms to real-life scenarios. 
In this paper, the criteria to assess are the capacity to overcome specific types of drifts. This section presents some benchmark scenarios that evaluate the ability to overcome some particular kinds of drifts. We also provide real-life application examples that would need assessment of a given benchmark.}


\subsection{Incremental Learning:}

\modif{Incremental learning scenarios assess the capacity of algorithms to deal with virtual concept drifts.}
Indeed, in a wide part of continual learning literature, incremental learning is associated with the \say{disjoint class scenarios} (\classesDrifts) \citep{Belouadah2018DeeSIL, wu2019large, douillard2020podnet}. One of the specificities of incremental learning is the addition of different concepts (classes, objects, actions) into a single model. Hence, the output space of the learning model grows through time.

\modif{Incremental learning can be formalized as follow:}

\modif{We consider two successive contexts $c_1$ and $c_2$ with $\Omega_{c_1}$ and $\Omega_{c_1}$ their respective set of concepts. Incremental learning can be formalized as:}
\begin{equation}
    \forall (x_1,y_1) \sim P(x,y|c_1), \forall (x_2,y_2) \sim P(x,y|c_2), ~ x_1 \neq x_2 \land y_1 \neq y_2.
    \label{eq:incremental}
\end{equation}
\modif{It means that concepts from different contexts are disjoint and that a drift in $c$ will lead to a data distribution with new concepts and without previously seen concepts.}

\textbf{Specifications:} 
Context drifts happen in the labels and observations, $\Omega_{c_1} \cap \Omega_{c_2} = \varnothing$, criterion are invariant \modif{and there is no real concept drift.}


In incremental learning, the learning model has to learn to differentiate data from different contexts. Hence, the learned function $f_{\theta}(\cdot)$ must respect:
\begin{equation}
    \forall (x_1,y_1) \sim P(x,y|c_1), \forall (x_2,y_2) \sim P(x,y|c_2), ~ f_{\theta}(x_1) \neq f_{\theta}(x_2).
    \label{eq:incremental_learning}
\end{equation}
%
%
 In this case, the goal is to incrementally train a model with the output space that grow through time.
 This is typically the case of a trained classification model that should learn a new class without forgetting the previous ones.
 From a practical point of view, it consists in getting different output for different tasks and preferably to be able to make predictions without knowing tasks label.
 In reinforcement learning different outputs of the model correspond to different actions, then incremental learning in reinforcement learning is about learning tasks that use different set of actions. For example, because in some tasks, some actions are invalid \citep{huang2020closer}, [\href{https://boring-guy.sh/posts/masking-rl/}{masking-rl-blogpost}].
 %
\modif{This scenario assesses the ability to learn or discriminate concepts that are not available simultaneously.}

\medskip

\underline{Example:} \textbf{Adding classes to a classifier}
\textit{A classifier is pretrained to recognized all known species of frogs. However, a researcher discovers a new species. This researcher would like to train the classifier to classify this new species correctly, but it does not have all the other species data. Then, he/she will need an incremental learning method to improve the classifier without forgetting.}

\medskip

\textbf{Note on supervised incremental learning:} In supervised incremental scenarios, drifts bring new classes. Hence, class labels are sufficient to detect new tasks and task labels are not necessary to detect drifts. Nevertheless, at test time, the availability of task index might still be required, e.g., to choose the right head in a multi-head architecture.

\subsection{Lifelong Learning}

\textit{Lifelong learning} \citep{Thrun95, Chen2018Lifelong} has often been use as a synonym of continual learning, however, we propose to associate it to a mode specific research field where a skills should be learn through a life-long task. 
\modif{This scenario assesses the ability to deal with} \instanceDrift  (cf section \ref{sec:drifts}) \citep{gepperth2016incremental, van2019three} where the concepts to learn in a task stays the same but the stream of data bring new data instances.
We consider it as a synonym to \textit{Never-Ending learning} \citep{Carlson10, Mitchell15}.

We formalize \textit{Lifelong learning} as follow:

We consider two context $c_1$ and $c_2$, with $\Omega_{c_1} \cap \Omega_{c_2}$ the set of concepts of context $c_1$ and $c_2$:
\begin{equation}
   \forall (x_2,y_2) \sim P(x,y|c_2), \exists (x_1,y_1) \sim P(x,y|c_1), ~ x_1 \neq x_2 \land y_1 = y_2.
\label{eq:lifelong}
\end{equation}
\textbf{Specifications:} The observation distribution changes through time, $\Omega_{c_2} \subseteq  \Omega_{c_1}$, criterion are invariant \modif{and there is no real concept drift.}

\medskip

\modif{Lifelong learning scenarios assess the ability to learn the similarity between data from different contexts.} Hence, the function $f_{\theta}(\cdot)$ to learn must respect:
\begin{equation}
     \forall (x_2,y_2) \sim P(x,y|c_2), \exists (x_1,y_1) \sim P(x,y|c_1), ~ f_{\theta}(x_1) = f_{\theta}(x_2).
    \label{eq:lifelong_learning}
\end{equation}
The output space of $f_{\theta}(\cdot)$ is fixed.

\medskip

In Lifelong learning, the agent's goal is to always improve at understanding concepts and generalizing them.
In classification, it would be learning on the same set of classes with more and more data of each task. 
In reinforcement learning, lifelong learning consists of improving at using a set of actions in more and more contexts with fixed reward function. 
It could be essential for applications where smart agents need to keep improving and gain efficiency from experiences on a specific task or set of tasks.

\medskip

\underline{Example:} \textbf{Predicting Weather} 
\textit{A model is trained to predict the weather. Hence, it should understand the dynamic of the weather and predict future weather from current observations. However, with global warming, this dynamic changes. Therefore, the algorithms need to adapt to this new dynamic to make good predictions with this new context (warmer global temperature). The model will need a lifelong learning approach to improve on the same task with different contexts and observations.}

\subsection{Learning under real-concept drifts}
Learning in a sequence of tasks where the labels change for a given sample has been less explored in continual learning recent literature, but many research studied it in the online learning literature \citep{hoi2018online}.

In a learning scenario, real concept drifts can be written as:
\begin{equation}
\forall (x_2,y_2) \sim P(x,y|c_2), \exists (x_1, y_1) \sim P(x,y|c_1), ~
 x_1 = x_2 \land y_1 \neq y_2.
    \label{eq:real_concept_drift}
\end{equation}

As explained in Sec \ref{sub:drifts}, real-concept drifts and criterion drift are similar in appearance and they can be treated similarly.

Real-concept drifts in classification scenarios are quite unusual since labels rarely change through time. However, in detection or segmentation, such drifts might happen when objects are annotated first as background and later as a specific object \citep{cermelli2020modeling,douillard2020plop}. 
The real-concept drifts might also happen if labels get more precise through time \citep{abdelsalam2021iirc}.
Real-concept drift scenarios might be approached in two ways, either we consider that the current label is the true one, and previous ones should be forgotten, either we consider than both have values and everything worth remembering.

In reinforcement learning, learning under real concept drift can happen when there is a change in the reward function.
\modif{A scenario with real concept drift assesses the capacity of a learning model to adapt to a change of ground truth. However, depending on the needs, the model could have to remember past ground truth or forget it.}

\textbf{Specifications:} The label/reward for a given observation/action changes through time but there is no drift in $p(x)$.

\medskip

\underline{Example:} \textbf{Diseases terminology update} 
 \textit{A model is trained to recognized diseases from a set of symptoms observations. However, the research community regularly update the terminology of diseases. For example, it might consider different two diseases that were previously considered as the same.
 Therefore, the model needs to adapt to make predictions compatible with this new terminology. The data potentially stayed the same, but the labels changed. We are in a real-concept drift scenario.}

\subsection{\modif{Summary}}

\modif{We summarize the different types of drifts described here and the scenarios that make possible to assess how continual learning algorithms can deal with them in the following list: }

\begin{itemize}
    \item \modif{Virtual Concept Drifts: Incremental learning scenario}
    \item \modif{Domain Drifts: Lifelong learning scenario}
    \item \modif{Real Concept Drifts (or criterion drifts): Learning under real-concept drift scenario}
\end{itemize}

Those three scenarios make a clear separation of the different challenges of continual learning. They can also be combined into hybrid scenarios. 
However, addressing them independently will can provide a clear set of quality for a given quality. \modif{One future work would be to estimate how transferable results from benchmarks are to real-life scenarios. A good benchmark provides a clear estimation of quality transferable to real-life applications.} 
It is important to note that for a fair comparison between scenarios they should be used with the same drift annotation policy and under the same assumptions, as described in sections \ref{sec:assumptions} and \ref{sub:drift_annotation}.

\section{Discussion}
\label{sec:discuss}

Continual learning is a vast field that gathered researchers who (1) strive to develop algorithms that learn as animals or humans do in similar environments and others  (2) who aim to create algorithms that can simply accumulate knowledge incrementally. 
In any case, continual learning is about learning under context drifts. 
In this section, we discuss the algorithms' autonomy level, the promises and limitations of meta-learning for continual learning, and the importance of distinguishing benchmarks from use cases to experiment algorithms performances.

\par \textbf{Autonomy and Indirect Supervision: } 
To address real-life scenario problems, algorithms need external supervision signals. This supervision might be the labels/rewards but it can also be indirect supervision given by the designer. For example, the way that data are selected and pre-processed for learning and validation can be seen as a supervision signal.  The model selection is generally also an external intervention.
In continual learning, the model may also have some insight of future tasks, such as the number of classes or difficulty \citep{Rahaf2019selfless,Douillard2020Insights}.
More generally any external intervention from the designer can be seen as supervision.

On the other hand, indirect supervision might be necessary for the algorithm to identify the task to solve. Hence, before aiming for a fully autonomous system or unsupervised learning, it might be interesting to differentiate supervision essential to understand the task from the supervision that simply eases learning.

The problem of autonomy in continual learning is often mentioned, and many approaches aim at fully unsupervised continual algorithms. 
However, a fully unsupervised learning algorithm can not identify the task to solve or the change in tasks and, therefore, can not solve any continual learning setting. Even if the reduction of supervision in continual learning is interesting and valuable, it should be done carefully to not aim at impossible problems.

\par \textbf{Meta-Learning: } 
In recent years, meta-learning \citep{hospedales2020meta} has been used in many continual-learning approaches \citep{riemer2018learning,Vuorio18,rajasegaran2020itaml,Javed2019Meta, liuincremental} to solve their perilous sequence of tasks. Meta-learning is a way to learn to continually learn. For this, meta-algorithms are trained on several continual tasks (support tasks) and tested on other continual tasks (test tasks). Meta-learning has also been used in continual learning for fast remembering an old task \citep{He2019TaskAC,caccia2020online}.

In the general case, meta-learning is a way to learn to learn by using support tasks to optimize parameters and/or hyper-parameters for future tasks. For example, it might find the best initialization for solving a future task. Meta-learning is a powerful framework to solve difficult learning tasks.
Nevertheless, the success of meta-learning depends on the similarity between support and test tasks. 

In the case of continual learning, if we suppose that meta-continual learning is used to train an algorithm for future (unknown) continual tasks. 
The similarity between training and test tasks is assumed from the beginning. If the future tasks are not as expected, the algorithm might fail to solve the tasks.
Unfortunately, it is not clear what kind of \say{similarity} between the train and test tasks is needed to make algorithms able to generalize to new continual tasks.
In continual learning, it might be related to the kind of drifts in the support tasks and the test tasks or to assumptions that hold in both cases.

\par \textbf{Benchmarks vs Use Cases: } 
Algorithms evaluations are supported by benchmarks.
%
Those benchmarks are often simplified scenarios targetting a  set of specific criterions. 
This simplicity of benchmarks is justified because it helps for a clearer evaluation. For example, the non-backtracking chain helps to evaluate the capacity of models to remember.
Unfortunately, without careful use, data and assumptions can be overfitted easily and produce misleading results.
One should then stay vigilant to use benchmarks properly to not exploit irrelevant assumptions or knowledge. 

Recent discussions in the continual learning community pointed out that current benchmarks are too limited and are not realistic enough to look like real-life scenarios. 
However, the goal of benchmarks is not to look like real-life use cases but to evaluate properly the set of criterions that would be useful for real-life applications. 
%
%
%
Then real-life use cases should not replace benchmarks for evaluation. Use cases can be used as a proof of concept or help to design the good criterion to maximize but the results they provide are not as valuable as those from a carefully designed benchmark.

Hence, benchmarks should be selected primarily for their evaluation quality and users should stay vigilant when maximizing performance to not overfit benchmarks data and assumptions. 
The creation of a well-designed dataset for continual learning remains an important challenge for the community notably because they are many types of application  scenarios and many data types \citep{douillard2021continuum}.

\section{Conclusion}

Continual learning is about learning in non-iid settings. In those settings, the data distribution may drift in various ways which disrupt the learning process.

In this paper, we start from the hypothesis that different drift types can not be processed in the same way by learning algorithms. Therefore, we describe the various kind of drifts and propose to use them to design a detailed description of continual-learning scenarios. 
The scenarios proposed in this paper are mainly characterized by their context drifts (data distribution drifts). However, they are meant to be used with a clear drift annotations policy, and a well-defined set of assumptions. 
The goal is to use well-characterized scenarios to evaluate continual learning algorithms and to measure how suited they are to deal with specific types of drifts under specific conditions.

Our description method should be a step toward a better understanding of benchmarks and the algorithm's capacity.
We believe that a better understanding of benchmarks will lead to easier transfer into real-life applications. 

\newpage

\bibliography{continual,others}
\bibliographystyle{unsrtnat}


\end{document}